\documentclass{article}

\usepackage{microtype}
\usepackage{graphicx}
\usepackage{subcaption}
\usepackage{booktabs} 

\usepackage{hyperref}

\usepackage{algorithm}
\usepackage{algorithmic}

\usepackage[preprint]{icml2026}

\usepackage{amsmath}
\usepackage{amssymb}
\usepackage{mathtools}
\usepackage{amsthm}
\usepackage{multirow}
\usepackage{booktabs}
\usepackage{makecell}
\usepackage{array}
\usepackage{arydshln} 
\usepackage{tabularx}

\usepackage[capitalize,noabbrev]{cleveref}

\theoremstyle{plain}

\theoremstyle{definition}

\theoremstyle{remark}

\newcommand{\method}{DualSpec}

\usepackage[textsize=tiny]{todonotes}

\icmltitlerunning{DualSpec: Accelerating Deep Research Agents
via Dual-Process Action Speculation}

\begin{document}

\twocolumn[
  \icmltitle{DualSpec: Accelerating Deep Research Agents\\ via Dual-Process Action Speculation}



  \icmlsetsymbol{equal}{*}
  \icmlsetsymbol{intern}{$\dagger$}

  \begin{icmlauthorlist}
    \icmlauthor{Shuzhang Zhong}{pku,intern}
    \icmlauthor{Baotong Lu}{msr,equal}
    \icmlauthor{Qi Chen}{msr}
    \icmlauthor{Chuanjie Liu}{ms}
    \icmlauthor{Fan Yang}{msr}
    \icmlauthor{Meng Li}{pku,equal}
  \end{icmlauthorlist}

  \icmlaffiliation{pku}{Peking University, Beijing, China}
  \icmlaffiliation{msr}{Microsoft Research}
  \icmlaffiliation{ms}{Microsoft}

  \icmlcorrespondingauthor{Baotong Lu}{baotonglu@microsoft.com}
  \icmlcorrespondingauthor{Meng Li}{meng.li@pku.edu.cn}

  \icmlkeywords{Machine Learning, ICML}

  \vskip 0.3in
]



\printAffiliationsAndNotice{$\dagger$Work performed during the internship while at Microsoft Research.} 

\begin{abstract}



Large language model-based deep research agents have been increasingly popular for addressing long-horizon information-seeking tasks, but they often incur high end-to-end latency due to extensive reasoning and frequent tool use. Speculation frameworks aim to reduce latency by overlapping action execution with reasoning; however, existing approaches typically rely on uniform speculation strategies and strict action matching, which limits inference speedups and robustness.

In this work, we revisit the speculate-verify paradigm for deep research agents through the lens of action heterogeneity. We show that \textit{Search} and \textit{Visit} actions exhibit fundamentally different reasoning and model capacity requirements: entropy-based analysis reveals that Search decisions have higher uncertainty and benefit significantly from explicit reasoning, whereas Visit decisions have lower entropy and depend primarily on model capacity. Motivated by this dual-process characteristic, we propose \method{}, a heterogeneous speculation framework equipped with a lightweight, confidence-based semantic verifier. Experiments across multiple models and benchmarks demonstrate that \method{} achieves up to 3.28$\times$ end-to-end speedup while maintaining accuracy comparable to fully reasoning agents.

\end{abstract}

\section{Introduction}

The increasing reasoning capabilities of large language models (LLMs) enable interaction with external tools and environments, driving the development of intelligent agents~\cite{yao2022react,shinn2023reflexion,schick2023toolformer}.
Among these, deep research agents have emerged as a prominent application for addressing open-ended, long-horizon research tasks with high information-seeking and reasoning demands~\cite{deepresearch}.
By iteratively reasoning and invoking external tools such as search engines, these agents accumulate evidence and refine hypotheses, extending beyond static question answering to complex research problems.

\begin{figure}
    \centering
    \includegraphics[width=\linewidth]{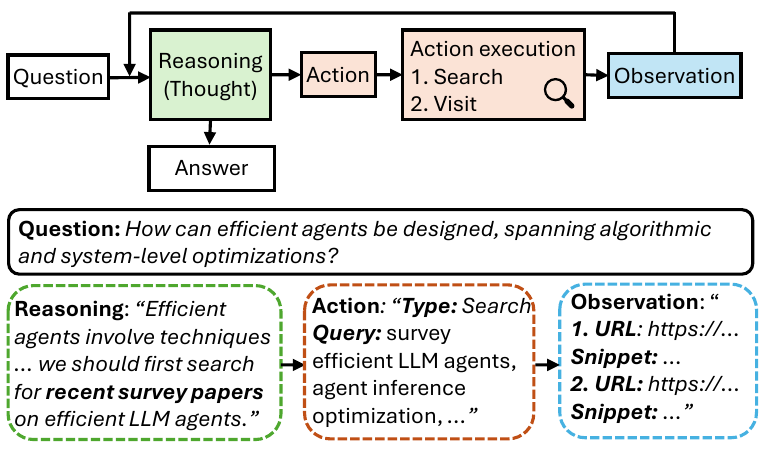}
    \caption{Deep research agent workflow. Deep research agents follow a Reason-Action-Observation loop, where the agent alternates between generating reasoning traces and executing actions (Search or Visit) to gather information.}
    \label{fig:deep_research}
\end{figure}


Despite their effectiveness, deep research agents often incur high inference latency.
As shown in Figure~\ref{fig:deep_research}, agents typically follow the ReAct paradigm~\cite{yao2022react} with strict sequential dependencies: the model must complete a reasoning trace before emitting an action and then wait for the resulting observation before proceeding.
Both reasoning and action execution can be time-consuming, particularly when using large models with long reasoning traces and external tools with variable response times\cite{kim2023llm,zhang2025optimizing}.
This reasoning--action--observation cycle repeats over many turns until a final answer is produced, often requiring minutes or longer for a single query.


A promising approach to reduce latency is the speculate--verify paradigm, where a lightweight model or strategy speculates the next action and executes it immediately, while the base model concurrently performs its reasoning~\cite{huang2025reducing,guan2025dynamic}.
If the base model's action matches the speculative one, the speculative observation is directly accepted, saving execution time; otherwise, the base model executes the action as usual.
Unlike speculative decoding~\cite{leviathan2023fast}, this approach operates at the action level rather than the token level, enabling parallelism between reasoning and tool use.
However, designing effective speculation and verification remains challenging: inaccurate speculation or conservative verification leads to frequent fallbacks and limited speedups, whereas overly permissive verification risks degrading agent performance.


In this work, we rethink speculate--verify for deep research agents through a principled analysis of action heterogeneity and verification trade-offs.
Existing lightweight speculation methods generally adopt either (i) small models with explicit reasoning or (ii) large models that emit actions without reasoning.
We observe that different action types exhibit distinct uncertainty profiles and thus require different speculation strategies.
Deep research agents primarily use two actions: \textit{Search}, which formulates a query to retrieve relevant webpages, and \textit{Visit}, which selects and accesses a specific URL from a candidate set.
Search involves high uncertainty in query formulation and benefits from strong reasoning, whereas Visit operates over a constrained action space and relies mainly on parametric knowledge.


We validate this distinction via end-to-end empirical evaluations, together with an entropy-based analysis of action decisions with and without reasoning.
Across settings, Search actions exhibit much higher uncertainty than Visit actions; explicit reasoning helps reduce uncertainty for Search but provides marginal gains for Visit.
This pattern aligns with the cognitive science distinction between \textit{System~2} (deliberative) and \textit{System~1} (intuitive) reasoning, with Search corresponding to the former and Visit to the latter.
Guided by these insights, we show that matching speculation strategies to action characteristics -- using a small reasoning model for Search and a large model without reasoning for Visit -- significantly improves speculation accuracy.


Verification is also critical for achieving efficiency without sacrificing performance.
Exact action matching is often overly restrictive, as semantically equivalent actions, especially queries, may differ at the token level.
Moreover, action-based verification typically requires the base model to complete reasoning before verification, placing reasoning on the critical path and limiting latency reductions.


Based on these observations, we propose \method, a heterogeneous action speculation framework for deep research agents that tailors speculation and verification to action-specific properties.
\method~uses a small reasoning model to speculate actions, while allowing the base model to concurrently generate a Visit action by skipping reasoning.
It dynamically selects an appropriate draft based on the reasoning state.
For verification, \method~leverages the base model's internal confidence rather than explicit action matching, removing base-model reasoning from the critical path while preserving the agent performance.


We implement \method~and evaluate it on two representative reasoning models, MiroThinker~\cite{team2025mirothinker} and Qwen-3~\cite{team2025tongyi}, using popular deep research benchmarks including GAIA-Text-103~\cite{wu2025webdancer}, XBench-DeepSearch~\cite{chen2025xbench}, and Seal-0~\cite{pham2025sealqa}.
\method~achieves up to \textbf{3.28}$\times$ end-to-end latency speedup while maintaining performance comparable to the fully reasoning base model.

\section{Background and Related Work}

\subsection{Deep Research Agents}
Given an input question, deep research agents operate in a multi-step loop that alternates between reasoning to generate an action, executing the tool call, and incorporating the response into its context, until producing a final answer~\cite{zhang2025agentorchestra,huang2025deep}.

\begin{figure}
\centering
\includegraphics[width=\linewidth]{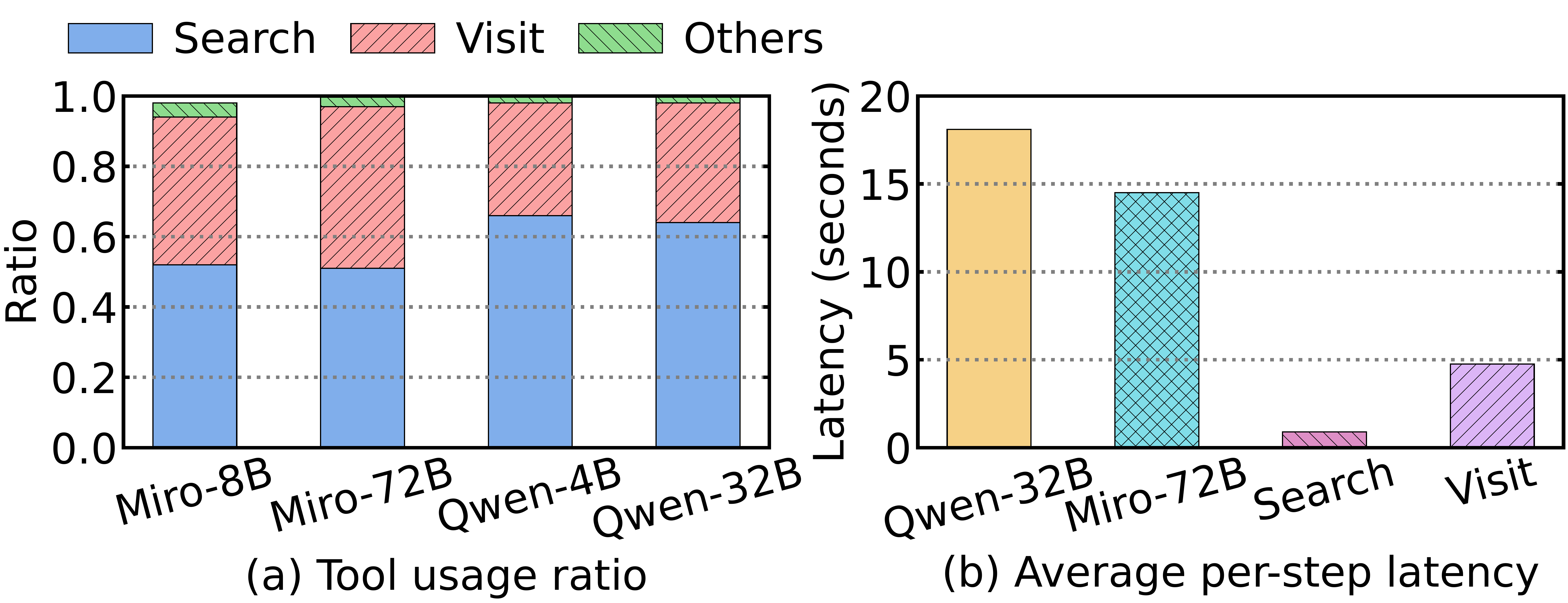}
\caption{
Deep research inference characteristics using different models. ``Miro'' denotes ``MiroThinker'' while 
``Qwen'' denotes ``Qwen-3''.     
(a) Tool usage ratio on the GAIA benchmark\footnotemark. (b) Time breakdown per step on model reasoning and tool execution. Model reasoning accounts for a significant fraction of the total latency.}
\label{fig:action_ratio}
\end{figure}

\footnotetext{Unless otherwise specified, all micro-level analyses are conducted on the GAIA benchmark; consistent trends are observed across other benchmarks.}


Most deep research agents rely on two core actions: Search and Visit.
Search consists of a query used to retrieve candidate webpages with brief snippets, while Visit selects a URL and specifies an instruction for extracting relevant information.
During execution,
Search directly queries a search engine, whereas Visit accesses the webpage and typically invokes an LLM to summarize task-relevant content according to the instruction\cite{nakano2021webgpt, zhou2023webarena}.
This design filters irrelevant information and limits unnecessary context growth.
Figure~\ref{fig:action_ratio}(a) shows that these actions occur at comparable frequencies, with Search used slightly more often due to query reformulation when results are unsatisfactory.
Other tool calls (e.g., code execution) constitute only a small fraction of steps.


Despite strong problem-solving performance, deep research agents often incur high latency due to their multi-step reasoning and tool-use workflows.
Figure~\ref{fig:action_ratio}(b) reports the per-step time breakdown measured on the A100 GPU.
Model reasoning dominates total latency, while tool execution introduces additional, though smaller, overheads.
Accumulated over many iterations, these costs lead to long end-to-end response times, limiting usability and deployment.

\subsection{Agent Optimization}


Stronger backbone models and richer tool interactions improve agent performance on challenging tasks~\cite{shang2025rstar2}, but often reduce time efficiency due to longer reasoning traces and more frequent tool use.
Recent work observes that agent steps vary in difficulty~\cite{zhang2023ecoassistant,saha2024system}, motivating approaches that delegate simpler steps to lightweight models while reserving complex reasoning for stronger models.


Speculate--verify paradigms have emerged as an effective strategy to reduce agent latency~\cite{ye2025speculative,guan2025dynamic,hua2024interactive,wang2025accelerating}.
Dynamic Speculative Planning~\cite{guan2025dynamic} employs a small reasoning model to draft an action and obtain its result while a stronger model performs full reasoning; the speculative action is verified against the base model's action using criteria such as minimum edit distance, and its tool response is reused upon agreement.
SPAgent~\cite{huang2025reducing} skips explicit reasoning in early stages and transitions to a speculate--verify phase later to maintain performance.

At a finer granularity, speculative decoding~\cite{leviathan2023fast} accelerates LLM inference at the token level by predicting future tokens with a smaller model and verifying them with a larger model. 
This approach is complementary to agent-level speculation and can be combined with it to further improve efficiency. 
SpecReason~\cite{pan2025specreason} reduces reasoning overhead by dynamically offloading simpler reasoning steps to a smaller model; however, it is not designed for agent-based settings with iterative tool use.


\subsection{Dual-Process Theory in LLMs}


Dual-process theory~\cite{chaiken1999dual} from cognitive science distinguishes between System~1, which is fast and intuitive, and System~2, which supports slower, deliberate reasoning.
This framework has recently been applied to interpret and guide LLM reasoning.
The System-1.x Planner~\cite{saha2024system} decomposes tasks into simpler and more complex sub-steps, assigning System~1 strategies to the former and System~2 strategies to the latter to improve efficiency.
However, it targets specific planning settings and requires extensive training.
How to systematically leverage dual-process principles in LLM-based agents with tool use, such as deep research agents, remains largely unexplored.
\section{Rethinking Speculate--Verify for Deep Research Agents}
\label{sec:motivation}

Existing speculate--verify frameworks typically apply a \emph{uniform} speculation strategy across all actions, either by (i) reducing reasoning depth (e.g., skipping explicit reasoning) or (ii) reducing model capacity (e.g., using a smaller speculator). While effective in some settings, this overlooks a key property of deep research agents: \emph{actions exhibit heterogeneous reasoning demands}. In this section, we provide empirical evidence that Search and Visit actions differ fundamentally in their sensitivity to reasoning depth and model capacity, motivating an action-aware speculation design. We further analyze verification trade-offs, highlighting the need to move beyond action matching.

\subsection{Speculation Under Action Heterogeneity}
We first examine the reasoning demands of different actions and then study how various speculation strategies affect the accuracy of drafting Search and Visit actions.

\begin{figure}
    \centering
    \includegraphics[width=\linewidth]{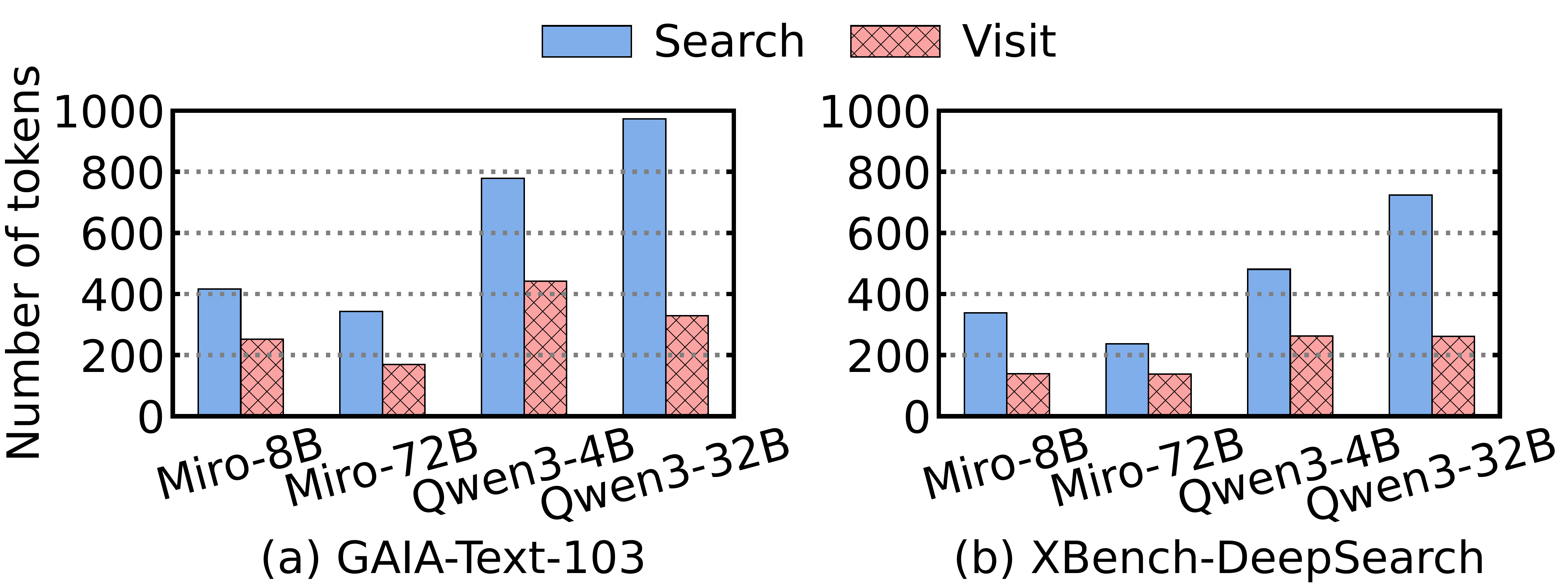}
    \caption{Average reasoning length for generating Search and Visit actions across models and benchmarks. Search requires significantly longer reasoning than Visit.}
    \label{fig:motivation_tokens}
\end{figure}

\paragraph{Observation 1: Search actions involve longer reasoning traces than Visit.} 
Figure~\ref{fig:motivation_tokens} shows the average reasoning length before emitting an action. Across all models, Search consistently requires $1.65$--$2.95$$\times$ more tokens than Visit, indicating that query formulation inherently demands more deliberation than webpage selection. Therefore, Search actions carry stronger reasoning requirements.



\begin{figure}
    \centering
    \includegraphics[width=\linewidth]{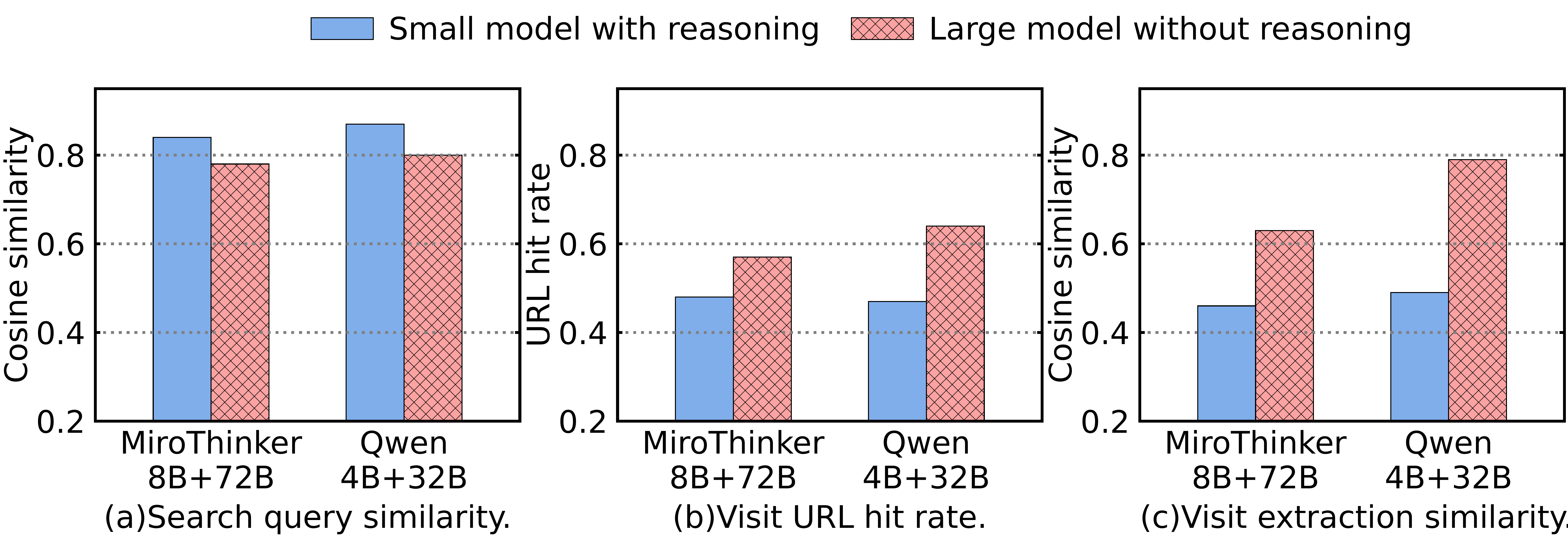}
    \caption{Action alignment comparison of two speculative methods relative to the Oracle (large model with reasoning) when drafting Search and Visit. (a) The small reasoning model produces  queries more aligned with the Oracle than the large model without reasoning. (b--c) For Visit, the large model skipping reasoning achieves higher accuracy in both URL selection and extraction instruction.}
    \label{fig:motivation_alignment}
\end{figure}

\paragraph{Observation 2: Effective speculation strategies differ across actions.}
We measure the alignment of speculative actions with an Oracle agent (large model with full reasoning) using two representative strategies: a small language model (SLM) with explicit reasoning and a large language model (LLM) that skips reasoning.
As shown in Figure~\ref{fig:motivation_alignment}(a), for Search actions, the SLM with reasoning consistently produces queries more aligned with the Oracle than the LLM without reasoning, measured via embedding-based cosine similarity~\cite{reimers-2019-sentence-bert}. This indicates that explicit reasoning is critical for query quality even under reduced model capacity. In contrast, Figure~\ref{fig:motivation_alignment}(b--c) shows that for Visit actions, the LLM without reasoning aligns more closely with the Oracle in both URL selection and extraction instruction, suggesting that deliberative reasoning is less essential for Visit, where pattern-based selection benefits more from model capacity.

\begin{table}[t]
    \centering
    \caption{End-to-end agent performance (pass@1 accuracy) on GAIA under action-level replacements.
We selectively replace the generation strategy for Search or Visit using lightweight methods.}
    \label{tab:action}
    {\small
    \begin{tabularx}{\columnwidth}{Xcc}
        \toprule
        Configuration & MiroThinker & Qwen3 \\
        \midrule
        LLM with Reasoning        & \textbf{63.11} & \textbf{29.13} \\
        \hdashline
        SLM with Reasoning for Search     & \textbf{63.11} & \textbf{29.13} \\
        LLM without Reasoning for Search  & 57.28 & 23.30 \\
        SLM without Reasoning for Search  & 55.34 &  22.33 \\
        \hdashline
        SLM with Reasoning for Visit      & 56.31 & 22.33 \\
        LLM without Reasoning for Visit   & \textbf{64.08} & \textbf{28.16} \\
        SLM without Reasoning for Visit  & 54.37 & 24.27 \\
        \bottomrule
    \end{tabularx}
    }
\end{table}



To assess end-to-end effects, we perform action-level interventions that selectively replace Search or Visit generation with lightweight methods while keeping the rest of the pipeline unchanged. We also include another variant of speculation combining reduced capacity and reduced reasoning (SLM without reasoning).
Table~\ref{tab:action} shows that assigning an SLM with reasoning to \textit{Search} preserves overall accuracy, whereas using an LLM without reasoning degrades performance. Conversely, generating \textit{Visit} with an LLM without reasoning yields the best results, while alternative choices lead to significant accuracy drops. Although speculative actions do not perfectly align with the Oracle (Figure~\ref{fig:motivation_alignment}), 
end-to-end accuracy remains high because of the tolerance of approximation in reasoning models, as long as appropriate inference pathways are chosen per action.



\paragraph{Key insight: Search as System~2, Visit as System~1.}
These results reveal a clear action-level dichotomy. Viewed through dual-process theory, Search exhibits System~2 behavior, requiring deliberative reasoning to translate underspecified research goals into effective queries. Visit aligns with System~1 behavior, where selection and extraction primarily rely on fast, pattern-based recognition encoded in model parameters. This distinction provides a principled foundation for action-aware speculation.




\subsection{Verification Beyond Action Matching}
Speculation alone is insufficient for reliable speedups; verification is essential to prevent error propagation. Most existing methods verify speculative actions via exact or approximate matching with the base model output, but this has two limitations.
First, action equivalence is hard to define. Exact matching is overly restrictive, as semantically equivalent Search queries may differ token-wise, causing unnecessary rejection, while approximate matching often requires additional modules (e.g., embedding models) and threshold tuning.
Second, verification typically places the base model's full reasoning trace on the critical path, limiting latency gains. More aggressive designs~\cite{guan2025dynamic} allow multi-step speculation with parallel verification, but failures in the middle require rolling back all following speculative steps, wasting computation. Achieving better accuracy-efficiency trade-offs therefore requires verification strategies beyond simple action matching.

\section{Theoretical Analysis}
\label{sec:theory}



This section provides a theoretical explanation for the empirical observations in Section~\ref{sec:motivation}.
We analyze why different actions exhibit distinct sensitivities to explicit reasoning, leading to different optimal speculation strategies.

Our central claim is that Search and Visit actions differ in intrinsic decision uncertainty.
We formalize this intuition via an entropy-based analysis of action policies and show that Search actions benefit significantly more from reasoning-induced uncertainty reduction than Visit actions, explaining the empirically observed System~2 versus System~1 dichotomy.

\subsection{Preliminaries: Action Policies and Entropy}

At each step, an agent observes a state $s$ from the accumulated context and then samples an action $a$ 
from a policy $\pi(\cdot \mid s)$.
We denote the action spaces corresponding to Search and Visit by $\mathcal{A}_{\textsc{search}}$ and $\mathcal{A}_{\textsc{visit}}$, respectively.

A natural measure of decision uncertainty is the conditional entropy of the action policy:
\begin{equation}
H(\pi(\cdot \mid s)) \;=\; -\sum_{a \in \mathcal{A}} \pi(a \mid s)\log \pi(a \mid s).
\label{eq:policy_entropy}
\end{equation}
Lower entropy indicates a more confident and concentrated decision, whereas higher entropy reflects ambiguity among many plausible actions.

In deep research agents, however, actions are expressed as open-ended language strings (e.g., search queries or extraction instructions), rendering exact computation of~\eqref{eq:policy_entropy} intractable.
We therefore adopt a token-level proxy based on the negative log-likelihood of the realized action.
Specifically, for an action represented as a token sequence $a=(t_1,\ldots,t_n)$, we define the mean token-level entropy proxy:
\begin{equation}
\bar{H}(a \mid s) \;=\; \frac{1}{n}\sum_{i=1}^n \left(-\log p(t_i \mid s, t_{<i})\right),
\label{eq:token_entropy_proxy}
\end{equation}
where smaller $\bar{H}(a \mid s)$ (i.e., higher average token log probability) indicates lower decision uncertainty.

\subsection{Intrinsic Entropy Gap Between Search and Visit}

We first examine the baseline uncertainty of different actions when generated \emph{without} explicit reasoning.
Intuitively, Search actions map a broad and ambiguous intent to a concrete query, for which many formulations may be reasonable.
In contrast, Visit operates on retrieved candidates and localized content, significantly constraining the decision space.

\begin{figure}
    \centering
    \includegraphics[width=\linewidth]{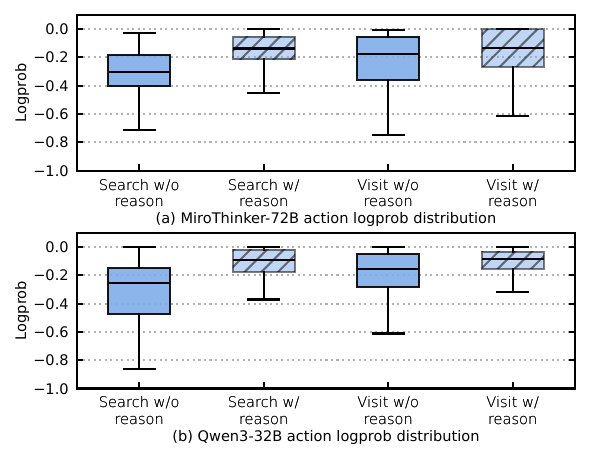}
    \caption{Action log probability distributions with and without reasoning. 
    A higher log probability indicates lower uncertainty.
    Without reasoning (dark blue color), Search actions exhibit lower log probabilities than Visit actions, indicating higher baseline decision uncertainty. 
    When reasoning is incorporated (light blue color with hatching), both action types see increased log probabilities, but the increase is significantly larger for Search actions, reflecting a greater reduction in uncertainty due to reasoning.}
    \label{fig:entropy_boxplot}
\end{figure}

This intuition is reflected in the following inequality:
\begin{equation}
    \mathbb{E}[\bar H(a\mid s)\mid a\in\mathcal{A}_{\textsc{search}}]
\;>\;
\mathbb{E}[\bar H(a\mid s)\mid a\in\mathcal{A}_{\textsc{visit}}].
\end{equation}
That is, under identical inference settings, Search actions exhibit higher average uncertainty than Visit actions.
Figure~\ref{fig:entropy_boxplot} visualizes this gap.
Each boxplot shows the distribution of mean token log probabilities for Search and Visit actions. 
When generated without reasoning, 
Search consistently exhibits lower log probabilities (higher $\bar{H}$) than Visit, suggesting less confident action distribution.

\subsection{Why Reasoning Helps: Entropy Reduction via Intermediate Structure}

We model explicit reasoning as the introduction of an intermediate latent variable $z$, corresponding to a reasoning trace that refines the decision context before the final action is generated.
This transforms the direct mapping $\pi(a \mid s)$ into a two-stage generation process:
\begin{equation}
\pi(a \mid s) \;=\; \sum_{z} \pi(z \mid s)\,\pi(a \mid s, z),
\label{eq:latent_factorization}
\end{equation}
where the final action is conditioned not only on the original state $s$, but also on the intermediate reasoning state $z$.
By a standard information-theoretic property~\cite{cover1999elements}, conditioning cannot increase entropy.
Formally,
\begin{equation}
\mathbb{E}_{z\sim \pi(\cdot\mid s)}\!\left[ H(\pi(\cdot \mid s, z)) \right] \;\le\; H(\pi(\cdot \mid s)).
\label{eq:entropy_reduction}
\end{equation}
Thus, access to a reasoning trace reduces action uncertainty and increases the likelihood of the realized action.

As shown in Figure~\ref{fig:entropy_boxplot}, incorporating reasoning consistently increases log probabilities, corresponding to a reduction in action-level uncertainty.
This reduction is most pronounced when the target decision relies on non-local associations that are not directly specified in the immediate input~\cite{prystawski2023think}.
Therefore, for Search, reasoning decomposes a global, underspecified mapping into a sequence of more localized sub-decisions, yielding a large reduction in uncertainty.
In contrast, Visit actions already exhibit low baseline entropy due to strong grounding in retrieved content.
Therefore, conditioning on an reasoning trace yields only a marginal additional reduction in uncertainty.

Taken together, this analysis explains why Search aligns more closely with \emph{System~2} behavior, while Visit is closer to \emph{System~1} behavior in deep research agents.

\section{\method{} Design}

\subsection{Overview}

\begin{figure}
    \centering
    \includegraphics[width=\linewidth]{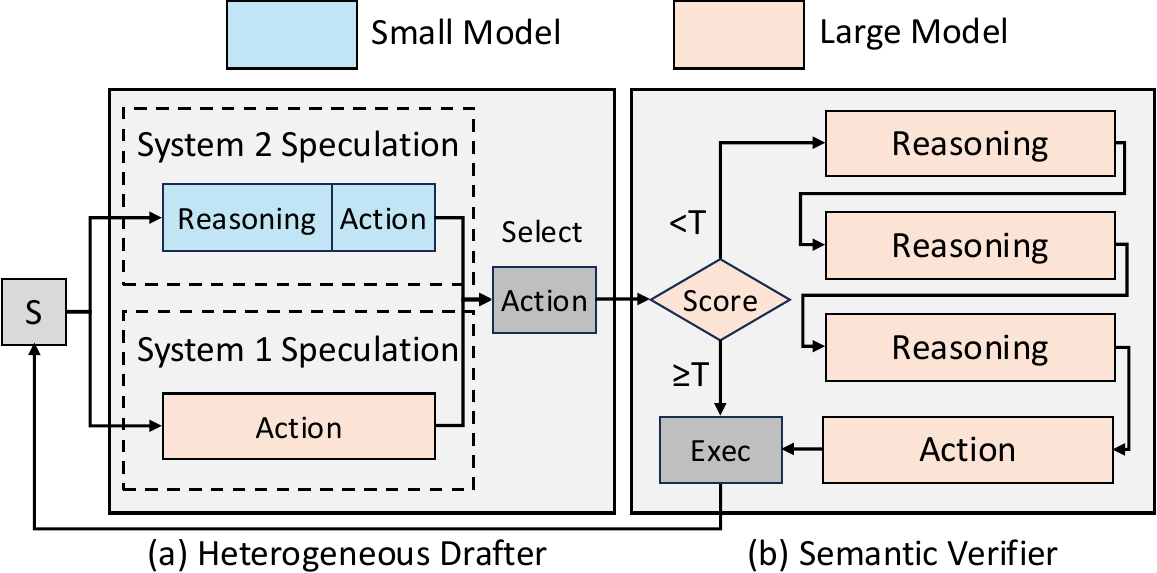}
    \caption{Overview of \method{}. 
     }
    \label{fig:overview}
\end{figure}

We propose \method{}, a dual-process speculative framework for deep research agents.
The core design principle of \method{} is to allocate inference resources \emph{heterogeneously} across actions for high speculation accuracy, while preserving end-to-end agent performance through lightweight, step-wise semantic verification.

As illustrated in Figure~\ref{fig:overview}, \method{} follows a draft--verify workflow.
At each decision step, \method{} generates two candidate actions in parallel:
(i) a \textit{System~2} draft produced by a small model with explicit reasoning, and
(ii) a \textit{System~1} draft via a large model skipping reasoning.
The framework then selects a provisional draft based on the action type and the reasoning footprint from the small-model output.
Finally, the selected draft is evaluated by a semantic verifier using a full-capacity base model.
Drafts judged to be semantically consistent with the current reasoning trajectory are accepted and executed directly; otherwise, \method{} falls back to full-capacity reasoning to regenerate the action.



\subsection{Heterogeneous Draft}
\label{subsec:hetero_draft}


\method{} implements heterogeneous drafting by producing two candidate actions at each step and adaptively selecting the one that best matches the inference demand of the current decision.
Formally, given the current state $s_t$, we generate a System~2 draft $(z_s, a_s)$ using SLM and
a System~1 draft $a_l$ using LLM.
The key question is how to select the final drafted action while retaining reasoning information that is valuable for long-horizon planning.

\paragraph{Action-aware selection.}
We use the action type predicted by the small-model draft as the primary routing signal.
If the small model generates a Search action, we retain $a_s$ as the draft action, as Search typically benefits from explicit reasoning and can be reliably handled by a smaller model when paired with a reasoning trace.
If the small model proposes a Visit action, we instead select the large-model draft $a_l$ in most cases, since Visit actions rely more heavily on the large model's parametric capacity to make direct decisions on concrete inputs.

\paragraph{Preserving long-horizon reasoning.}
An important exception arises when the small-model draft produces a long reasoning trace before emitting its action.
Empirically, such extended reasoning often contains global analysis or intermediate summaries that remain useful beyond the current step, regardless of whether the final action is Search or Visit.
To preserve this information, when the length of the small-model reasoning exceeds a threshold $\tau_{\text{think}}$, we choose the full draft $(z_s, a_s)$ even if the action type is Visit.
This mechanism ensures that high-level reasoning is not discarded when it may benefit subsequent decisions.

\subsection{Semantic Verification}
\label{subsec:semantic_verification}

To maintain end-to-end accuracy, \method{} performs lightweight semantic verification at every step.
Instead of enforcing action-level matching, the verifier assesses whether the drafted reasoning and action are likely to make meaningful progress.
This design is motivated by the observation that intermediate agent decisions are often tolerant to approximation as illustrated in table~\ref{tab:action}.
Moreover, this approach avoids the reasoning delays from the base models to generate actions, enabling faster verification.   

Given the current state $s_t$ and a draft consisting of an optional reasoning trace $z_t$ and a candidate action $a_t$, we query the large model as a critic and ask it to answer \texttt{Yes} or \texttt{No}.
The prompt instructs the critic to jointly assess (i) whether the reasoning is coherent (if present) and (ii) whether the proposed action is useful for making progress.
The exact prompt template is provided in Appendix~\ref{app:verifier_prompt}.

Although the critic produces a discrete verdict, a \emph{continuous} signal is necessary to trade off speed and reliability.
We therefore convert the critic's output distribution into a real-valued confidence score.

Let $p_{\mathrm{acc}}(s_t,z_t,a_t)$ and $p_{\mathrm{rej}}(s_t,z_t,a_t)$ denote the critic's probabilities of answering \texttt{Yes} and \texttt{No}, respectively.
We define the verification score as the log-probability margin:
\begin{equation}
\mathrm{score}(s_t, z_t, a_t)
=
\log p_{\mathrm{acc}}(s_t, z_t, a_t)
-
\log p_{\mathrm{rej}}(s_t, z_t, a_t),
\label{eq:verifier_score}
\end{equation}
which corresponds to the log-odds of acceptance and provides a stable, monotonic measure of verifier confidence.

We accept the draft if its score exceeds a threshold $\tau$:
\begin{equation}
\textsc{Accept}(z_t, a_t)
\quad \text{if} \quad
\mathrm{score}(s_t, z_t, a_t) \ge \tau,
\label{eq:accept_rule}
\end{equation}
and otherwise trigger fallback.
Fallback regenerates the step using the full-capacity model with explicit reasoning and continues execution with the regenerated action.
This verification-and-fallback procedure follows a standard speculative pattern: propose a fast approximate step, validate it with a stronger critic, and only pay the cost of full reasoning when the draft is unlikely to be reliable.

Since the score scale depends on the critic model, we select $\tau$ offline on a held-out development set.
We sweep candidate thresholds and choose a fixed $\tau$ that preserves end-to-end accuracy while maximizing the acceptance rate, and keep it fixed at runtime.
This allows \method{} to allocate expensive full-capacity reasoning only to the minority of uncertain steps, improving overall time-to-solution. Additional empirical evidence on the effectiveness of this verifier score is provided in Appendix~\ref{app:verifier_scores}.

\begin{figure*}[h]
    \centering
    \includegraphics[width=0.9\linewidth]{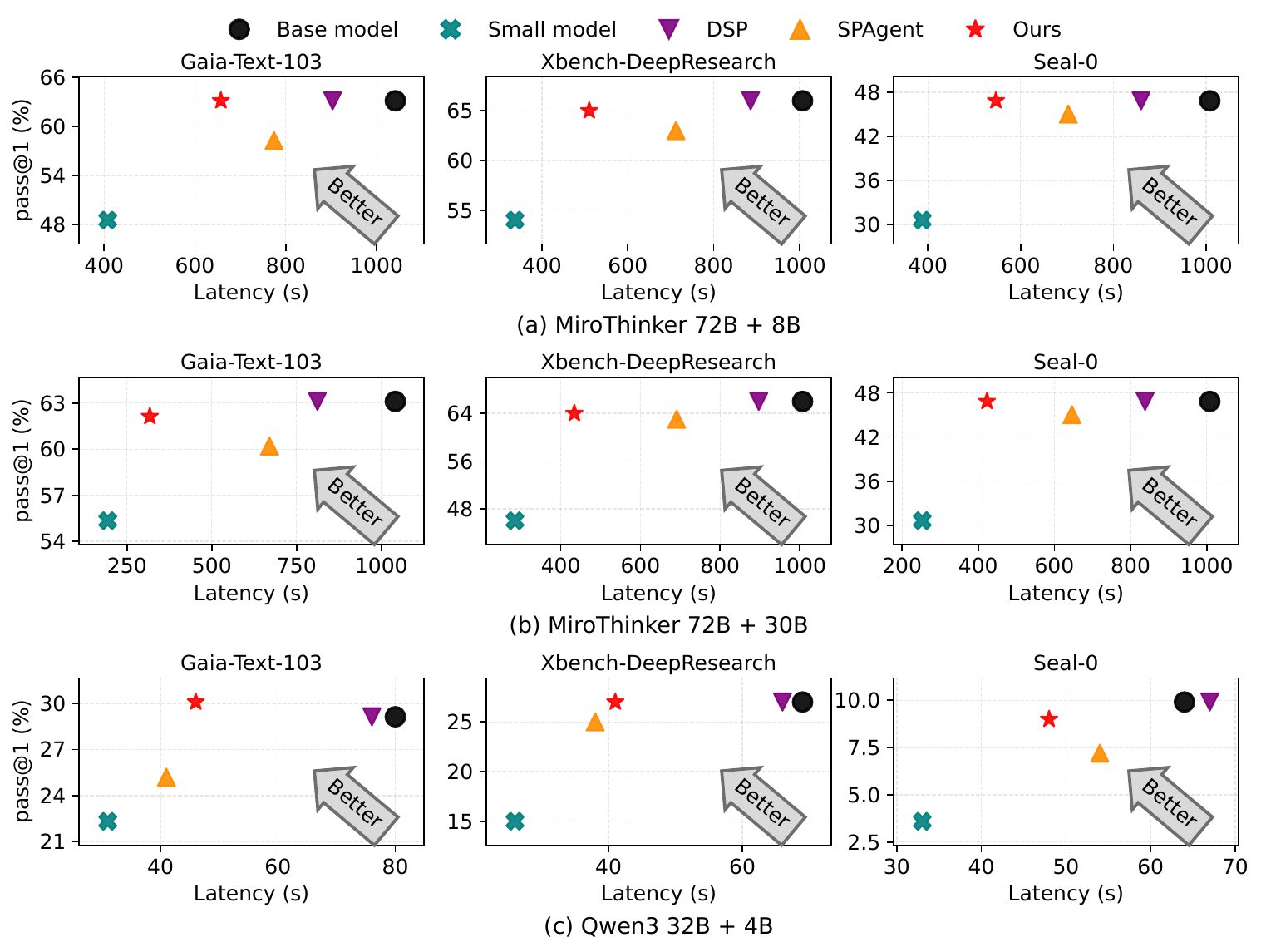}
    \caption{Comparison of the accuracy (pass@1) and latency of different schemes across model combinations.
\method{} consistently reduces end-to-end latency by \textbf{1.33–3.28$\times$} (\(\sim\)2$\times$ on average) over the base model while maintaining comparable accuracy.
Compared with DSP and SPAgent, \method{} achieves a better accuracy–latency trade-off across all datasets and model pairs.}
    \label{fig:main_results}
\end{figure*}

\section{Experiments}

\subsection{Experimental Setup}

\paragraph{Models.}
We evaluate \method{} under three dual-model configurations: 
MiroThinker-v1.0-72B + MiroThinker-v1.0-8B, 
MiroThinker-v1.0-72B + MiroThinker-v1.0-30B-A3B, and 
Qwen3-32B + Qwen3-4B~\cite{team2025mirothinker,yang2025qwen3}. All models are deployed single-tenant with one \textsc{NVIDIA} A100 GPU per model and batch\_size=$4$ for inference. The MiroThinker Models were quantized to 4-bit for inference, while the Qwen Models use native FP8 quantization, to avoid out-of-memory in GPUs.

\paragraph{Datasets.}
Experiments are conducted on three representative deep-research benchmarks:
including GAIA-Text-103~\cite{wu2025webdancer}, XBench-DeepSearch~\cite{chen2025xbench} and Seal-0~\cite{pham2025sealqa}.

\paragraph{Frameworks.}

We build our agent on the MiroMind deep‑research framework.
Tool invocation follows the MCP (Model Context Protocol) interface to standardize tool signatures and I/O, ensuring consistent argument formatting and result parsing across models and datasets.
Within this setup, Search calls are executed via the Bing API, and Visit operations (page fetching and readable content extraction) are served by Jina, providing a fixed backend for web querying and page processing throughout all experiments. The models are serving under SGLang\cite{zheng2024sglang}.

\paragraph{Baselines.}

We compare with two speculative agent frameworks: DSP~\cite{guan2025dynamic} and SPAgent~\cite{huang2025reducing}.
DSP targets planning tasks and accepts a draft only if it matches the base action (minimum edit distance),
while SPAgent is tailored to web search, skipping verification early and enforcing strict action matching later.
In contrast to their uniform drafting and action-alignment verification, \method{} uses heterogeneous drafting for System~2/System~1 actions and semantic verification that accepts trajectory-consistent drafts without exact action equivalence.

\paragraph{Verifier threshold.}
We tune the verifier threshold $\tau$ on a held-out split of GAIA, targeting an intervention rate of $\sim$20\%,
and reuse the same $\tau$ in our experiments.

\subsection{Main Results}

Figure~\ref{fig:main_results} reports end-to-end \emph{latency} vs.\ \emph{pass@1} across three model pairs and three deep‑research benchmarks. 
Overall, \method{} attains \textbf{1.33–3.28$\times$} speedup over the base model, averaging \textbf{$\sim$2$\times$}, while maintaining comparable pass@1.
Across datasets and model combinations, the points for \method{} consistently move left (lower latency) with negligible accuracy degradation, indicating a better accuracy–latency operating point than uniform speculative baselines.

By analyzing each model pair, we observe 1.8$\times$ acceleration on MiroThinker‑72B + 8B, 2.6$\times$ on MiroThinker‑72B + 30B, and 1.5$\times$ on Qwen3‑32B + 4B.
The larger gain with the 30B configuration arises from its MoE design that each forward activates roughly \emph{3B} parameters; at the same time, its stronger base capability reduces the number of base model interventions, further reducing end‑to‑end time.
Consequently, using 30B‑A3B as the base model delivers the highest overall speedup among the evaluated pairs.

\subsection{Ablation Studies}

\subsubsection{speculation methods}

\begin{table}[h]
  \centering
  \caption{Performance under different speculation schemes.}
  \label{tab:spec_ablation}
  \setlength{\tabcolsep}{4pt}
  \begin{tabular}{ccccc}
    \toprule
    \textbf{Models} & \textbf{Datasets} & \textbf{Speculation} & \textbf{Acc} & \textbf{Lat} \\
    \midrule
    \multirow{8}{*}{\makecell[c]{Miro\\72B+8B}} 
      & \multirow{4}{*}{GAIA}
        & Origin         & 63.1 & 1041 \\
      & & LLM w/o Reason & 59.2 & 575 \\
      & & SLM w/ Reason  & 56.3 & 651 \\
      & & Heterogeneous   & 63.1 & 605 \\
      \cmidrule(lr){2-5}
      & \multirow{4}{*}{Xbench}
        & Origin         & 66    & 1007 \\
      & & LLM w/o Reason & 65    & 492 \\
      & & SLM w/ Reason  & 65    & 501 \\
      & & Heterogeneous   & 66    & 480 \\
    \midrule
    \multirow{8}{*}{\makecell[c]{Qwen\\32B+4B}} 
      & \multirow{4}{*}{GAIA}
        & Origin         & 29.1 & 80 \\
      & & LLM w/o Reason & 27.1  & 67 \\
      & & SLM w/ Reason  & 25.2  & 32 \\
      & & Heterogeneous   & 30.1 & 46 \\
      \cmidrule(lr){2-5}
      & \multirow{4}{*}{Xbench}
        & Origin         & 27    & 69 \\
      & & LLM w/o Reason & 25    & 46 \\
      & & SLM w/ Reason  & 26    & 49 \\
      & & Heterogeneous   & 27    & 41 \\
    \bottomrule
  \end{tabular}
\end{table}

To analyze the impact of heterogeneous speculation on performance, we fix the verification setting and vary only the speculation strategy, comparing our heterogeneous approach against the small-model-only methods and skipping-reasoning methods.
As shown in Table~\ref{tab:spec_ablation}, heterogeneous speculation consistently achieves a better accuracy–latency balance than either \textit{LLM w/o Reason} or \textit{SLM w/ Reason}, maintaining accuracy while reducing end-to-end latency across model pairs and datasets.

\subsubsection{Intervention Rate}

\begin{figure}
    \centering
    \includegraphics[width=\linewidth]{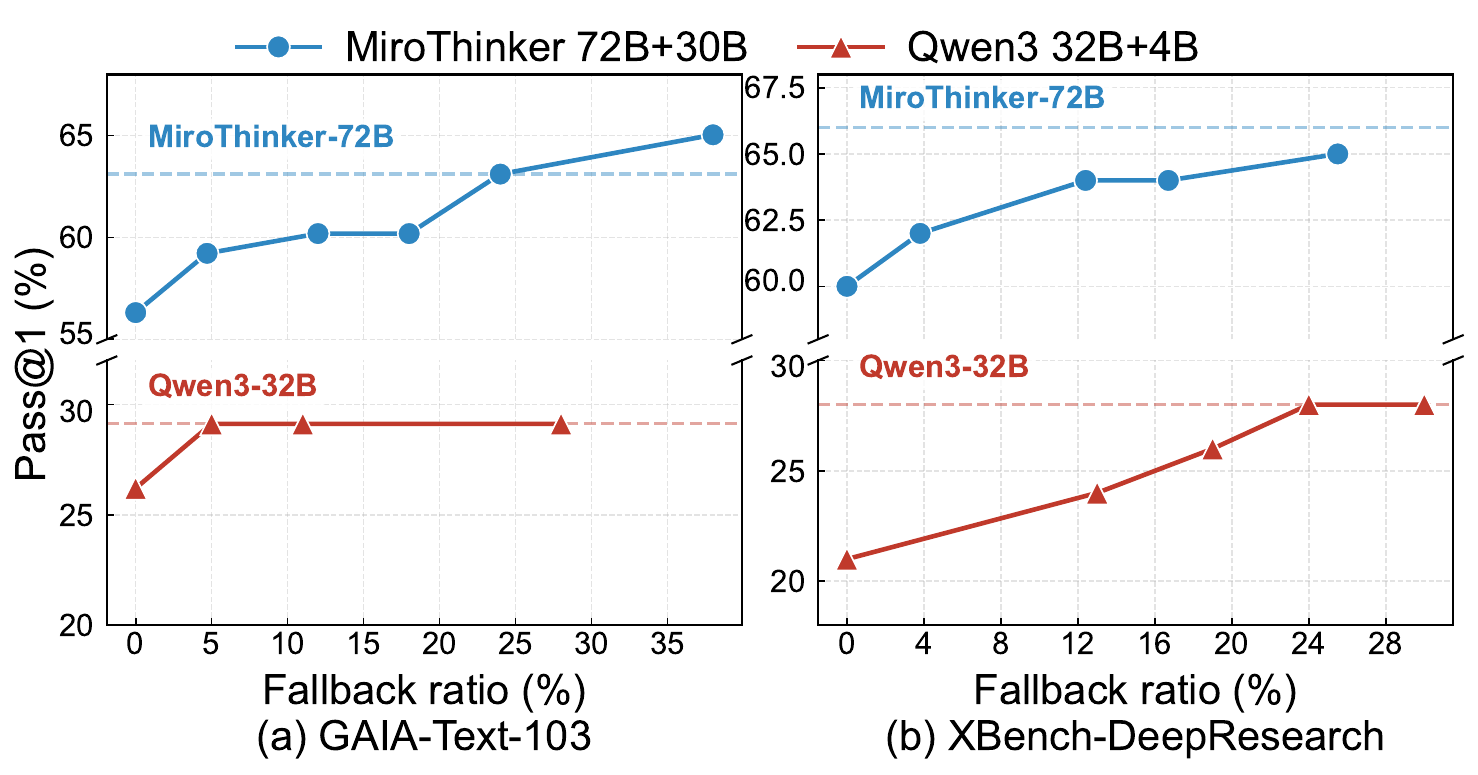}
    \caption{Accuracy (pass@1) as a function of the reasoning intervention rate under a fixed drafting policy.}
    \label{fig:fallback_ratio}
\end{figure}

We study how accuracy changes with the frequency of the large model intervention. With a fixed speculator, we vary the verifier threshold, which indirectly controls the intervention rate. Accuracy increases as the intervention rate rises and then saturates. In practice, we observe that an intervention rate of about \textbf{20\% to 30\%} already reaches accuracy comparable to the base model, while retaining most of the latency benefit of heterogeneous speculation.

This trend is consistent across model families and datasets, with minor shifts in the saturation point. While tighter thresholds further improve accuracy, the returns diminish once the rate reaches the low-to-mid twenties. We therefore tune the threshold to target an intervention rate near 20\%, recovering near-base accuracy without sacrificing the efficiency benefits of sparse large-model reasoning.

\section{Conclusion}

We introduce \method{}, an efficient framework that accelerates deep research agents through heterogeneous action speculation.
Our key insight is that actions exhibit different uncertainty levels: Search often requires deliberative reasoning, whereas Visit is typically more deterministic and can be executed without reasoning.
Exploiting this asymmetry, \method{} integrates action-specific draft policies with semantic verification to enable reliable speculative execution while removing large-model reasoning from the critical path.
Experiments across multiple benchmarks show that \method{} significantly reduces latency while maintaining strong task success rates, highlighting the importance of action-aware speculation for scalable agentic systems.

\section*{Impact Statement}
This paper presents work whose goal is to advance the field of Machine Learning. There are many potential societal consequences of our work, none which we feel must be specifically highlighted here.

\newpage
\bibliography{example_paper}
\bibliographystyle{icml2026}

\newpage
\appendix
\onecolumn

\section{Verifier Prompt and Additional Analysis}
\label{app:verifier}

\subsection{Verifier Prompt Template}
\label{app:verifier_prompt}

Given the current state $s_t$ and a draft output consisting of an optional reasoning trace $z_t$ and a candidate action $a_t$, we query the large model as a critic to output a binary judgment (\texttt{Yes} or \texttt{No}).
The critic is instructed to jointly assess (i) whether the trajectory is making new progress toward the user goal and (ii) whether the proposed action is grounded and useful.
When the draft pathway skips explicit reasoning, we set $z_t=\emptyset$.
The exact prompt used for the critic is shown below.

\begin{quote}
\small
\noindent\textbf{[SYSTEM: TRAJECTORY AUDIT]}\\
Review the recent steps (context). Is the agent making \textbf{NEW PROGRESS}?\\[0.3em]
\noindent\textbf{REJECT ("No") if:}
\begin{enumerate}
    \item \textbf{Stagnation}: Repeating similar queries or visiting same URLs (Looping).
    \item \textbf{Ungrounded Answer}: The Final Answer is NOT supported by the \textit{retrieved search results}.
    \item \textbf{Lazy/Drift}: Queries are nested, vague, or irrelevant to User's Goal.
\end{enumerate}
\noindent Verdict: Is the trajectory HEALTHY and PROGRESSING?\\
Answer only "Yes" or "No".
\end{quote}

\subsection{Verifier Score Distributions}
\label{app:verifier_scores}

To further validate that the verifier score provides a meaningful signal, we analyze the score distributions on complete trajectories.
Specifically, we apply two full-size critics, MiroThinker-72B and Qwen3-32B, to evaluate step-level draft outputs produced by MiroThinker-8B and Qwen3-4B, respectively.
We report two aggregated statistics per trajectory: the mean score across all steps (\texttt{Mean}) and the 25th percentile score (\texttt{p25}), where \texttt{p25} emphasizes low-confidence segments within a trajectory.

\begin{figure}[H]
    \centering
    \includegraphics[width=0.8\linewidth]{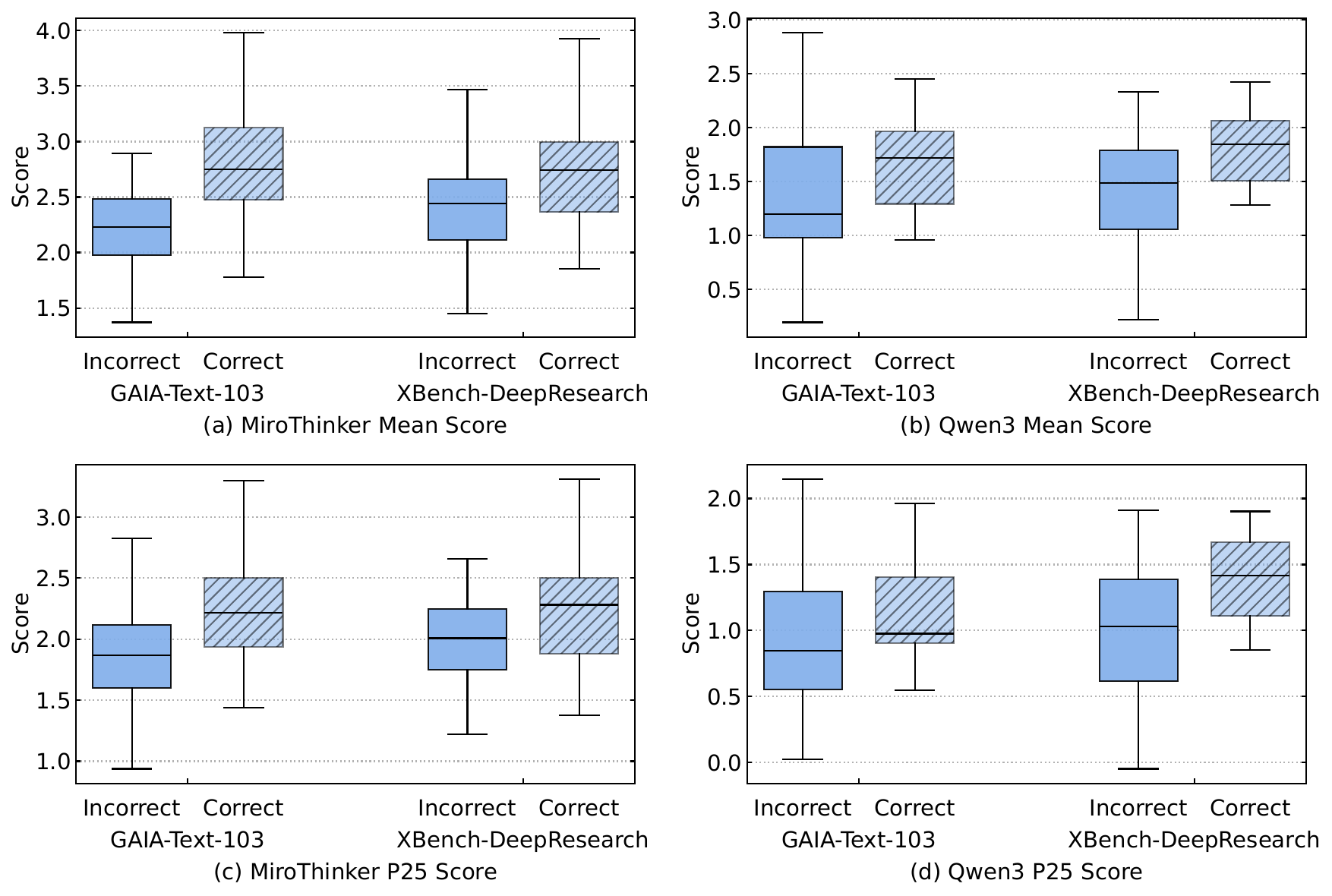}
    \caption{Verifier score distributions on GAIA and XBench-DeepResearch.
    We report trajectory-level aggregated verifier scores for (a) MiroThinker mean score, (b) Qwen mean score,
    (c) MiroThinker p25 score, and (d) Qwen p25 score.
    Correct trajectories consistently receive higher scores than incorrect ones, suggesting that the verifier provides a useful reliability signal.}
    \label{fig:verifier_score_dist}
\end{figure}

Figure~\ref{fig:verifier_score_dist} shows the score distributions on GAIA and XBench-DeepResearch, grouped by whether the final answer is correct.
Across both datasets and both model families, correct trajectories consistently exhibit higher verifier scores than incorrect ones under both \texttt{Mean} and \texttt{p25}.
This indicates that the verifier score correlates with end-to-end task success, supporting its use as a lightweight reliability signal for controlling speculative execution.

\end{document}